\documentclass[runningheads]{llncs}
\usepackage{graphicx}
\usepackage{array}
\usepackage{tabularx}
\newcolumntype{C}[1]{>{\raggedright\arraybackslash}p{#1}}
\usepackage{multicol}
\usepackage{multirow}
\usepackage{epigraph}
\begin{document}
\title{Toward a Perspectivist Turn in Ground Truthing for Predictive Computing}
\titlerunning{Data Perspectivist turn}

\author{Valerio Basile\inst{1} \and
Federico Cabitza\inst{2} \and
Andrea Campagner\inst{2} \and
Michael Fell\inst{1}}

\authorrunning{V. Basile et al.}

\institute{Università di Torino, Turin, Italy %\email{valerio.basile@unito.it}
\\
\and
Università di Milano-Bicocca, Milan, Italy\\
%\email{federico.cabitza@unimib.it}
}

\maketitle       % typeset the header of the contribution

\begin{abstract}
Most Artificial Intelligence applications are based on supervised machine learning (ML), which ultimately grounds on manually annotated data. The annotation process is often performed in terms of a majority vote and this has been proved to be often problematic, as highlighted by recent studies on the evaluation of ML models.
In this article we describe and advocate for a different paradigm, which we call \emph{data perspectivism}, which moves away from traditional gold standard datasets, towards the adoption of methods that integrate the opinions and perspectives of the human subjects involved in the knowledge representation step of ML processes. Drawing on previous works which inspired our proposal
we describe the potential of our proposal for not only the more subjective tasks (e.g. those related to human language) but also to tasks commonly understood as objective (e.g. medical decision making), and present the main advantages of adopting a perspectivist stance in ML, as well as possible disadvantages, and various ways in which such a stance can be implemented in practice. Finally, we share a set of recommendations and outline a research agenda to advance the perspectivist stance in ML.

\end{abstract}

%FC The page limit for a full paper (where the abstract in this case needs to include no more than 250 words) is equal to 12 pages (excluding references).

\section{Motivations and Background}
\label{sec:intro}

%\epigraph{Don't confuse the truth with the opinion of the majority.}{Jean Cocteau}

%\epigraph{Truth is a sea of grass tossed by the wind.}{Elias Canetti}

Much of contemporary Artificial Intelligence is based on supervised machine learning, a methodology that leverages large datasets that are manually ``annotated'', that is, associated with categorical labels in a case-wise manner (see Table~\ref{table:glossary}). The common and undisputed pipeline for the preparation of these datasets includes the definition of the classification schema (i.e., the eligible labels), data collection and annotation\footnote{Since annotation, interpretation and rating are terms that can be used indiscriminately to indicate both the process (i.e. the labelling) and its effect or result (i.e., the labels), here we prefer to speak of phenomenon and objects (to be classified); which is associated with a judgement and some labels, as result of an interpretation act by some raters, although all the synonyms mentioned above are equally good as long as they do not generate notational ambiguity. In the same vein, we speak of rating and rater, where other authors could use the terms label and annotator, respectively. To help the reader with this terminological fuss, we produced Table~\ref{table:glossary}.}, and, critically, label aggregation. This latter step is often performed in terms of a majority vote. This has been proved to be problematic in many work settings, especially when the object to classify is so complex that most of the raters can get it wrong and the real experts are a minority~\cite{aroyo2015truth,basile2021end,beyer2020we,cabitza2020if,cabitza2019elephant}, or the object is ambiguous as it is often the case in Natural Language Processing (NLP) tasks \cite{artstein2008inter}. Recently, these issues have attracted the interest of the AI and ML community, thanks to initiatives like the Data Nutrition Project\footnote{\url{https://datanutrition.org/}} or the Data Statements~\cite{bender2018data}, as well as the recent proposal of Data-Centric AI by Andrew Ng\footnote{ \url{https://www.youtube.com/watch?v=06-AZXmwHjo}}, who argued for greater attention and transparency related to the data production and annotation processes.
In this article, we argue that the issues with aggregation are not necessarily only a \emph{data quality} problem (but rather also regard process quality and responsibility) and, furthermore, are not necessarily restricted to highly subjective phenomena, but are also relevant to fields where the phenomena to be classify are traditionally considered ``objective'', like in clinical decision support.
%(in Section~\ref{sec:2flavor} we will briefly challenge this trite and somehow misleading dichotomy)
.

\begin{table}[!htb]
\centering
\label{table:glossary}
\caption{Glossary of the main concepts discussed in this article.}
  \scriptsize
\begin{tabularx}{\textwidth}{|l|X|X|}
\hline
term &
  synonyms and related concepts &
  definition \\ \hline
annotation &
  labelling, interpretation, rating, labeling, judgment &
  Either the process or the result of labelling (i.e., associating labels with) a set of objects. \\ \hline
ground truth &
  gold standard, label set, target set, also dataset from which training, validation and test sets are extracted &
  set of the labels to be associated with the training objects. Ground truthing is the process where such a set is created. \\ \hline

label &
  category, class, target value, jugdement, code, symbol, token &
  symbol (or set of symbols) by which to annotate (label, classify) objects in the training set. It can be any term out of a list, a category, a code, a level of measurement on an ordinal scale, a value. \\ \hline
  model &
  algorithm &
  mathematical representation used to classify new objects and predict unknown outcomes on the basis of input unlabelled objects. The output of a learner. \\ \hline
  object &
  instance, case, phenomenon, record, row, example &
  symbolic representation of an object of the phenomenon or reality of interest to be properly classified or predicted by the ML model. \\ \hline
  
 phenomenon &
  object &
  what is represented as data in each record and is to be classified, either in the training set (by raters) or as output of the predictive model. \\ \hline
rater &
  annotator, observer, labeller &
  who is involved in labelling (i.e, ground truthing), that is who is supposed to classify training data to give as input to machine learning procedures. \\ \hline

reduction &
  derivation &
  the procedure of transforming a multi-rater (multi-label) ground truth into a single-label one (with only one label for each object, case), e.g., by majority voting. \\ \hline
  supervised (learning) &
  (learning) by example &
  type of machine learning where labeled dataset (i.e., training sets) is used to train models that classify data or predict outcomes. \\ \hline
  
\end{tabularx}%
\end{table}

We propose and argue for a paradigm shift, which could move away from monolithic, majority-aggregated gold standard datasets, towards the adoption of methods that more comprehensively and inclusively integrate the opinions and perspectives of the human subjects involved in the knowledge representation step of modeling processes. Our proposal comes with important and still-to-investigate implications: first, supervised models equipped with full, non-aggregated annotations have been reported to exhibit a better prediction capability \cite{akhtar2020modeling,campagner2021ground,sudre2019let}, in virtue of a better representation of the phenomena of interest; secondly, new techniques for AI explainability can be devised that describe the classifications of the model in terms of multiple and alternative (if not complementary) perspectives \cite{basile2021end,noble2012minority}; finally, we should consider the ethical implications of the above mentioned shift and its impact on cognitive computing, whereas the new generation of models can give voice to, and express, a diversity of perspectives, rather than being a mere reflection of the majority \cite{noble2012minority,rizos2020average}.

\section{Strong and weak data perspectivism}

\begin{figure}[bth]
  \centering
  \includegraphics[width=1\textwidth]{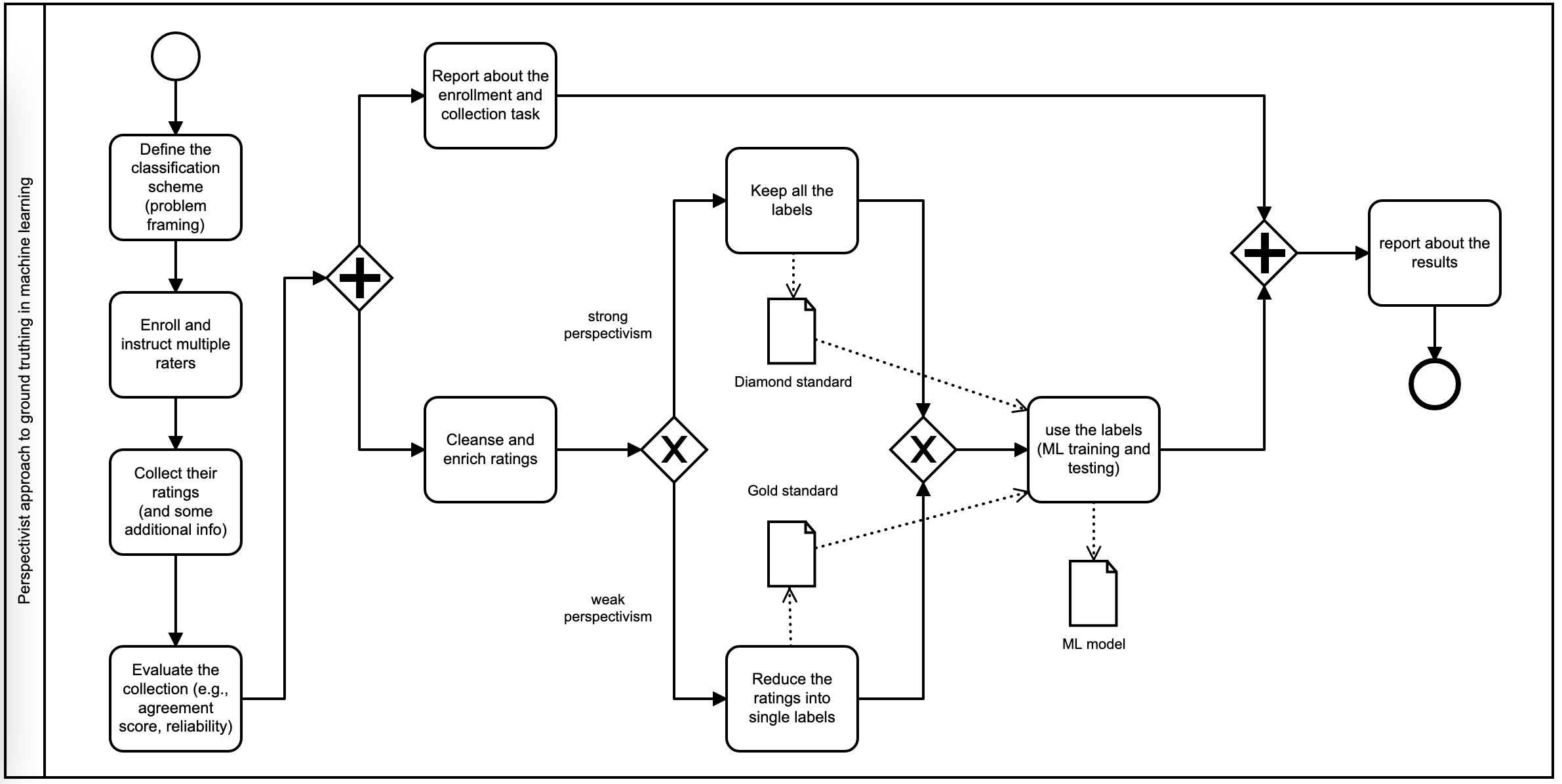}
  \caption{A BPMN (Business Process Model and Notation) diagram of the ground truthing process in a perspectivist setting. Many tasks are common with more traditional ML pipelines, but the core distinction with these latter ones lies in the exclusive gateway (X) in the centre of the diagram. Parallel gateways (+) indicate opportunities for parallel activities, which we made explicit to emphasize the importance of the comprehensive reporting of the ground truthing process.}
  \label{fig:process}
\end{figure}

As anticipated above, in this paper we propose what we denote as a \emph{perspectivist} approach in producing the ground truth (i.e., the ground truthing task) to be used in supervised classification tasks by systems developed with Machine Learning (ML) methods and techniques. 
This general stance can be articulated in two main versions, which could be connoted as either a \emph{weak} or \emph{strong} approach (see Figure~\ref{fig:process}).

A \emph{weak perspectivist} approach is adopted when researchers involved in ground truthing are not content to collect a single label for each object to be classified, that is to produce a \emph{gold standard} label set; but rather aim to collect as many raters \emph{and} annotations as possible, i.e., to build what in~\cite{campagner2021ground} has been called \emph{diamond standard} (see Figure~\ref{fig:process}). We emphasize the distinction between raters and annotations because, as simply as it can be, raters could express more than one label for a given object to classify \cite{dumitrache2018capturing,li2018multi} (also as a way to express their indecision in case of strictly alternative labels), or they could expressly be asked to rank available labels in terms of pertinence or to associate each class with a confidence/probability level \cite{chen2013pairwise}. 

One could rightly wonder why ML researchers would want to collect such redundant information about the phenomena about which they design and develop decision support systems that are usually aimed at improving human decision making by proposing the one best label for each object to classify \cite{aroyo2015truth}. For now, we are not going to dispute that conceiving the output of such systems as single labels is the best way to improve human decision making\footnote{As a matter of fact, alternative ways, like in conformal prediction, could improve it more~\cite{balasubramanian2014conformal}, especially if we consider improvement also beyond the mere dimension of error rate}; even when the output must be a single label, ML researchers could want to collect multiple labels for single objects to classify when they deem relying on single judgments either too \emph{limiting} (like in multi-label tasks, where labelling categories are not disjoint and they overlap to some extent) or too \emph{error-prone} \cite{vaughan2017making}. This could happen for two main reasons: 1) distrust in the raters involved, as it can be the case in crowdsourcing initiatives or questionnaire-driven studies where researchers cannot oversee the annotation task or ensure its accuracy \cite{eickhoff2013increasing,vaughan2017making}; 2) the recognized variability (what later we will call their \emph{inter-subjective} nature) of the phenomenon of interest~\cite{aroyo2015truth}, that is the recognition that different raters could classify the same object differently, and not necessarily because they are wrong. Indeed, ratings can differ not only because the raters are fallible, but also for a number of other factors, among which we recall: the intrinsic ambiguity and complexity of the phenomenon, including the so-called \emph{cumulative mess}, that is the condition when the same object can legitimately be classified as many things at the same time~\cite{bechmann2019unsupervised}; the (in)stability of the phenomenon over time~\cite{eickhoff2013increasing}; the complexity of the task, also in terms of number of distinct states or configurations of the phenomenon to detect~\cite{salminen2018inter} and of the concentration that is requested to the raters, and their necessary proficiency to detect and understand the phenomenon~\cite{jagabathula2017identifying}; the raters' susceptibility to somehow systematic cognitive biases, both at individual and team level, like overconfidence, confirmation and availability bias, anchoring and halo effects~\cite{eickhoff2018cognitive}, as well as to more contingent and context-driven external factors (what Kahneman and colleagues have denoted as \emph{chance variability of judgments}, or ``decision noise''~\cite{kahneman2016noise}).

Thus, collecting multiple labels from a sample of individuals can help ML researcher get a sample of perceptions, opinions and judgments that could be maximally \emph{representative} of the population of interest; also, and more practically speaking, doing so would help them draw from such a sample some \emph{qualified} majority \cite{cabitza2020if} for the sake of higher accuracy\footnote{With qualified majority we intend either a \emph{statistically significant} majority (with respect to some hypothesis testing procedure), or an \emph{overwhelming} majority, regardless of how this may be defined, and thus irrespective of what perspectives are considered and how they are distributed within the sample}. 

Weak perspectivism would then advocate the production of gold standards by considering and taking into account multiple perspectives. This does not necessarily require to make each perspective symbolically explicit, in terms of multiple labels: a meeting where multiple experts are invited to share their opinion about an object would be perspectivist as long as all people involved could express their opinions and views in a discussion, which could then be summarized into single positions. Likewise, a weak perspectivist approach would require to collect multiple labels from a corresponding number of observers or raters but then would combine these labels and select one single label for each object to be annotated, mostly by some kind of majority voting (e.g., weighted voting). This process has been called \emph{perspective reduction} in~\cite{campagner2021ground}.

On the other hand, we would speak of \emph{strong perspectivism} whenever the researchers' aim is to collect multiple labels, or multiple data about each class, about a specific object, and \emph{keep them} all in the subsequent phases of training or benchmarking of the classification models. Doing so certainly impacts model training and evaluation, but can be realized in several ways, of varying complexity \cite{campagner2021ground,sudre2019let,uma2020case}. The easiest way, that is the most backward-compatible way that does not require ad-hoc implementations, is to replicate each object in the training set to reflect the number of times this object has been associated with a certain label by the raters \cite{zhang2018ensemble}; nonetheless, also other methods have been proposed in the literature, that we will better describe in Section \ref{sec:RW}.

%FC dare la terminologia
%FC dobbiamo riportare i termini che riteniamo più giusti e rappresentativi e poi riportare i sinonimi in varie comuntà. Ad esempio:
%Rater o observer or annotator? The subject of the process.
%(Process of) Annotation, labelling, rating, or ground truthing?
%What is the object of this process? Object, 
%, case, phenomenon, as general terms to denote texts, images, set of codes and numbers that represent some object or phenomenon.

\section{Between objectivity and subjectivity}
\label{sec:2flavor}

From a conceptual point of view, the process of ground truthing can be understood as a process of collection of a set of \emph{measures}, according to the theory of scales of measurement~\cite{cicchetti2006rating}. In this light, any mapping from a phenomenon or object of interest (also a portion of the reality) to a sign (mostly a symbol of conventional meaning) can be assimilated to a measurement; the related measure (symbol) that can be used to annotate its data representation is but the result of the process above that can be called in multiple, yet equivalent, ways: rating, judgement, perception, opinion, annotation.

%Measures of inter-rater agreement as ways to quantitively assess observer variability.

As we said above, a perspectivist approach  can be advocated especially for the annotation of those phenomena that we would consider as ``highly subjective'', from a common sense stance, that is affected by high \emph{observer variability}, as measured with some inter-rater agreement metrics, like kappa~\cite{fleiss1971measuring}, alpha~\cite{krippendorff2018content} or rho~\cite{cabitza2020if}. 
According to a common definition, a judgment is considered \emph{subjective} when it is mainly ``based on, or influenced by, personal feelings, tastes, or opinions.''; we usually contrast this concept with that of \emph{objective}, a term that characterises judgements that, ideally, are not influenced by personal feelings or idiosyncrasies and which, on a practical level, the vast majority of people would see and label in the same way (barring clear errors or oversights). 

In metrology, the science of measurement, the above terms are associated with slightly different meanings: A recent proposal~\cite{maul2019intersubjectivity} asserts that any measurement that aims to be dependable (as ground truth must be, by definition) should exhibit both the characteristics of \emph{objectivity}, that is being related to an object, even to a ``non-physical property such as chess playing ability''; and that of being \emph{interpersonal}, that is independent of the subject or ``identically interpretable by different measurers''~\cite{maul2019intersubjectivity}.

\begin{figure}[bth]
 \centering
 \includegraphics[width=\textwidth]{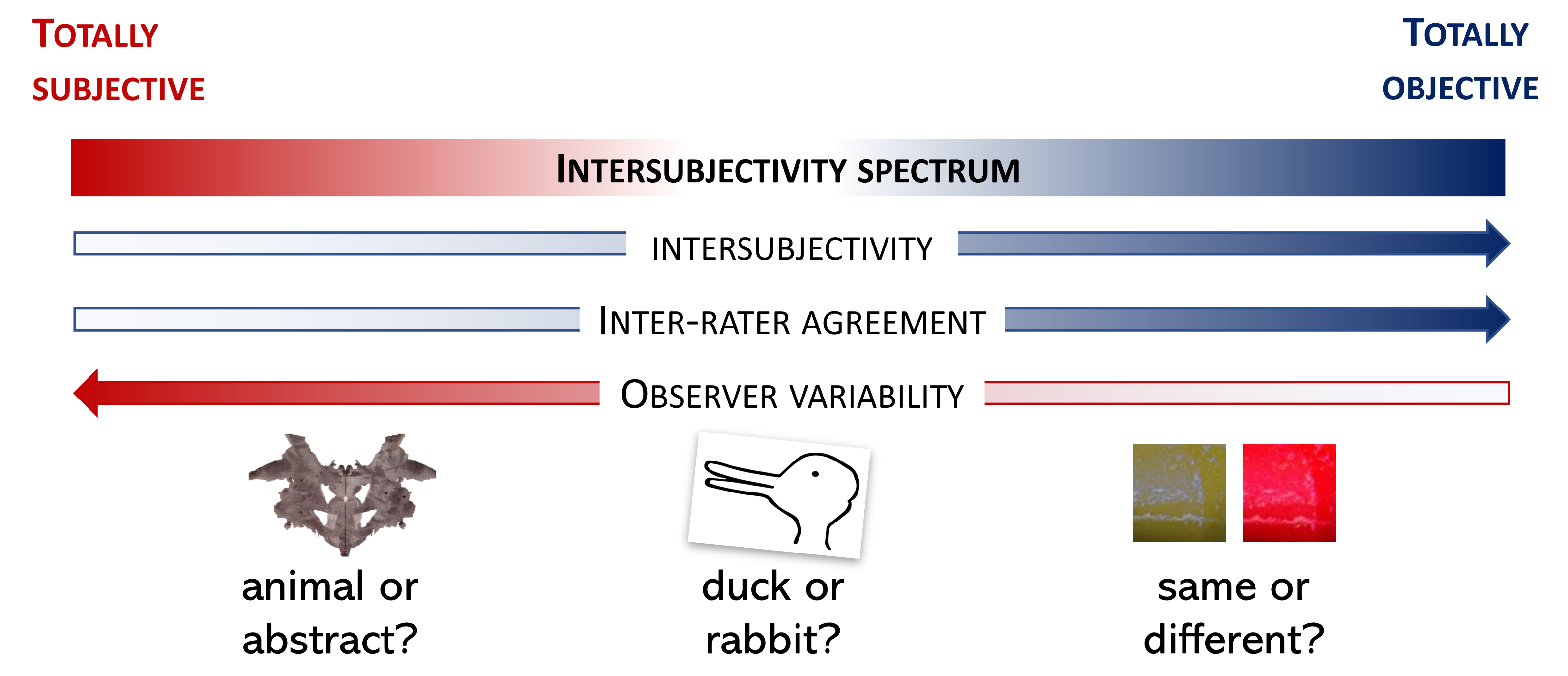}
 \caption{The intersubjectivity spectrum and its relationship with inter-rater agreement.}
  \label{fig:spectrum}
\end{figure}

In the perspectivist discourse, we recognize that the opposition obj/subj-ective is somehow too much affected by common sense and, worse yet, value judgments (in that subjective usually means ``bad'' or, to our practical aims, unreliable; and objective means good, and reliable). Therefore, we prefer to consider these two concepts as sort of ideal extremes of a full range of ways in which phenomena can present themselves to the interpretation and judgement of the members of a human community, what we call the \emph{intersubjectivity spectrum} (see Figure~\ref{fig:spectrum}). 

In this range, the extremes are essentially abstract: no phenomenon, and hence opinion on it, can be said totally subjective, as human beings are all immersed in a culture and social ensemble (trivially, they may share a language and semantic conventions), even if we recognized the ineffable and undetermined nature of the phenomenon at hand; likewise, full objectivity is a mirage because any judgement (or measurement) is situated and affected by contingent and contextual conditions, even if executed by a sensor or machine. 

Appraising where a certain phenomenon of interest can be located in this conceptual continuous range, that is assessing its level of intersubjectivity, can be preparatory to understand what kind of perspectivist approach, whether weak or strong, to adopt in one's ML pipeline; and whether keeping multiple labels allows to preserve individual expression and intuition or, conversely, can undermine the reliability (and hence utility) of the ground truth. To this respect, highly intersubjective phenomena would be phenomena that are close to the objective end of the intersubjective spectrum and for which disagreement among the raters involved is limited or absent. On the other hand, low-intersubjectivity objects would be those for which the complexity and ambiguity of the phenomenon at hand, but also the experience, commitment and competence that are required of the involved raters (like in medical diagnosis) would make one expect low agreement scores~\cite{salminen2018inter}. This can happen also in unsuspected domains, like medical diagnosis: just as a few examples, agreement in the 
%morphologic classifications of histotype 
grading of ovarian cancer slides has been found associated 
%with kappa scores of 60\%, and with much lower scores in grading ($\kapp=.25$)~\cite{}; 
with kappa scores lower than 30\% ($\kappa=.25$~\cite{barnard2018inter}), like in ischemia detection from EEGs ($\kappa=.29$~\cite{mehta2011agreement}) and in phase lag detection in ECGs ($\kappa=.23$~\cite{foreman2016generalized}). Such a low agreement is anything but rare in classification tasks (even in medicine); being aware of this can call for the application of some advanced methods, where the single contributions collected are weighted so as to identify the most reliable ones in the subsequent phases of ML development.

\section{Related work}
\label{sec:RW}

In what follows, we report on the main research contributions that either inspired or adopt some form of perspectivism in the ML pipeline (see Figure~\ref{fig:process}). Annotation is a crossing task that regard different types of data, including texts (and any part of them), images (including segmented portions), and more or less structured data, like case records.

The scientific communities interested in ML, like NLP, Computer Vision (CV) and medical informatics,
have traditionally relied on gold standard datasets to design, develop and evaluate supervised models: these datasets have usually been obtained by the annotation of a single rater or by means of the majority aggregation of few raters, and their reliability, and representativeness of the real task under consideration, have scarcely been questioned.

In the recent years, however, due to the huge increase in raw data availability, the increasing reliance on crowdsourcing and similar annotation protocols has highlighted the issue of observer variability in Machine Learning tasks \cite{aroyo2015truth,cabitza2019elephant}, an issue which was already well known in certain settings such as the computational linguistics \cite{artstein2008inter} and medical ones \cite{jewett1992potential,lok1998accuracy}: Beyer et al. \cite{beyer2020we} and Yun et al. \cite{yun2021re} showed the majority-aggregated labels in the original ImageNet dataset to be not representative of the images in the dataset due to observer variability and the un-reliability of the annotation process; Svensson et al \cite{svenssonautomated2015} noted the influence of observer variability on the performance of Machine Learning in a task of cancer detection and proposed ways to measure model performance in settings affected by variability; similarly, the impact of observer variability in NLP has been explored by Akhtar et al. \cite{akhtar2019polarization,akhtar2020modeling} in the task of hate speech detection.

While many works, and specifically those focusing on the crowdsourced learning setting, have adopted a weak perspectivist stance for the development of ML methods able to account for this observer variability \cite{chang2017revolt,heinecke2019crowdpac,schaekermann2018resolvable,tao2020label}; the need for ML methods explicitly taking into account a strong perspectivist approach, however, has only recently started to become a focus of research: Akhtar et al. \cite{akhtar2020modeling} showed that a strong
perspectivist approach to model training may also lead to performance increase; Sudre et al. \cite{sudre2019let} and Guan et al. \cite{guan2018said} proposed a multi-task approach to deal with observer variability in medical imaging, showing how jointly learning the majority consensus and the individual raters' labels improves classification accuracy; Sharmanska et al \cite{sharmanska2016ambiguity} showed how accounting for disagreements among raters may help improving the performance in CV tasks with crowdsourced labels; Peterson et al. \cite{peterson2019human} showed that accounting for raters' disagreement and uncertainty may lead to generalizability and performance improvements in CV tasks; Plank et al. \cite{plank2014learning} proposed a loss based on inter-rater agreement metrics to address variability in part-of-speech tagging; Uma et al. \cite{uma2020case} proposed the use of soft losses as a perspectivist approach for the training of ML models; Campagner et al \cite{campagner2021ground} proposed a soft loss ensemble learning method, inspired by possibility theory and three-way decisions, for the training of ML models in perspectivist settings.

In a similar direction, recent works~\cite{basile2021end,cabitza2020if,rizos2020average} explicitly explored the impact of strong perspectivism on the development and evaluation of supervised models, also from a more conceptual perspective. In particular, in \cite{basile2021end,cabitza2020if} experiments are presented in support of the thesis that disagreement in annotation may come from the subjectivity (i.e., low inter-subjectivity) of a task to varying extent \cite{salminen2018inter}, and therefore it should not be cast away as noise in the data, but rather it should be systematically accounted for at evaluation time.
Furthermore, in~\cite{basile2021end,rizos2020average} the authors also show an additional advantage of strong perspectivism in supervised learning, namely its potential impact on the interpretability and fairness of the models. In an experiment on real data (an annotated corpus of hate speech), the authors of~\cite{basile2021end} show how individual labels can be used to cluster the raters by affinity, leading to the emergence of patterns that helps identifying socio-demographic aspects of the raters themselves, which are in principle opaque, especially in a crowdsourcing scenario; while in~\cite{rizos2020average}, the authors describe how a similar approach could be used to detect biases in the data and labels provided by raters.  The related works, and the main ML techniques for perspectivist ground truths, are summarized in Table \ref{tab:methods}.

\begin{table}[!hbt]
  \centering
  \caption{A summary of methods proposed in the literature to train ML models based on perspectivist ground truths.}
  \label{tab:methods}
  \scriptsize
  \begin{tabular}{C{3cm}|c|c|C{6cm}}
     Method & Type & References & Description \\\hline
     Qualified Majority & Weak & \cite{cabitza2020if,kahneman2016noise} & Collect the annotations of at least 12 raters; if there is a qualified (i.e., statistically significant) majority keep the majority label, otherwise discard the object\\\hline
     Object Replication/Weighting & Weak & \cite{akhtar2019polarization,plank2014learning,sharmanska2016ambiguity,zhang2018ensemble} & Replicate or weight each object in the training set depending on the number of labels provided by the raters, or on some measure of disagreement\\\hline
     Consensus Labeling & Weak & \cite{chang2017revolt,schaekermann2018resolvable} & Use a participatory consensus building process (e.g. Delphi method) to annotate the objects\\\hline
     Weighted Majority & Weak & \cite{cabitza2020if,heinecke2019crowdpac,tao2020label} & Score the raters based on an estimate of their reliability; compute a weighted majority to annotate the object\\\hline
     Regularized Multi-task Learning & Strong & \cite{guan2018said,sudre2019let} & Keep all labels; train a model that jointly predicts the majority label and all the individual raters' labels\\\hline
     Ensemble Methods & Strong & \cite{akhtar2020modeling,campagner2021ground,svenssonautomated2015,zhang2018ensemble} & Sample multiple datasets from the original multi-rater one, according to some criteria, e.g., the distribution of labels, or the labelling rater; train an ensemble model encompassing several base models, one for each sampled dataset\\\hline
     Soft Loss Learning & Strong & \cite{campagner2021ground,peterson2019human,plank2014learning,uma2020case} & Associate each object in the training with a distribution, or set, over the labels provided by the raters; train a model using a modified loss function
  \end{tabular}
\end{table}

% Benchmann and Bowker ask: what are the political consequences of such subjective processes (labelling industries in the 2010s (e.g. Mechanical Turk) that are widely used to identify features and other labels in training data)?
% ground truths create problems if we turn to the classification theory of ubiquitous classification and cumulative mess
% 4 assumptions on ground truth; (4) future predictions will match the ground truth registered for past interactions [Florio 2020]
% definition of the task may be more narrowly interpreted by algorithms, not taking into consideration an ecological approach to the consequences on other tasks such as *societal inclusion* [Akhtar 2020] and social cohesion

\section{Looking at the two sides of the same coin}
\label{sec:2sides}

As anticipated above, a perspectivist approach to ground truthing requires to preserve the classification multiplicity instead of getting rid of it by majority voting (if original labels have been produced) or consensus surveys (if original labels do not exist). Obviously, as also highlighted in Section \ref{sec:RW} when discussing the related works setting the background for our proposal, this comes with some advantages and also some shortcomings, which we discuss in what follows.  

The main benefits of the perspectivist approach are: 
\begin{enumerate}
  \item To provide a theoretical backbone that recognizes and accepts the categorical irreducibility of some phenomena. This is especially relevant to those phenomena which exhibit a natural ambiguity, such as many tasks in NLP \cite{artstein2008inter}, or seemingly inconsistent clinical manifestations \cite{cabitza2019elephant};
  \item To extract valuable knowledge from what it is usually discarded as noise (cf. label noise \cite{kahneman2016noise}), i.e., disagreement. Such extra information is valuable for a decision support to be more useful in border-line and complex cases;
  \item To avoid to ratify and legitimize the opinion of raters belonging to a majority group, re-iterating their truth in seemingly objective advice. Instead, the perspectivist approach aims at giving voice to the few who hold a minority view \cite{noble2012minority}, or to those who are intimidated in collective debates;
  \item To be able to build models that learn typical human error patterns (if it is plausible to define ``errors'' on the basis of minority stances) and use this information as a form of decision support;
  \item To be able to develop models that can leverage label-uncertain \cite{uma2020case} and fuzzy data \cite{campagner2021ground} to improve performance, generalizability and robustness;
  \item To allow for three-way, fuzzy, probabilistic methods that represent the subjectivity and uncertainty in the considered phenomena \cite{cabitza2019elephant,kompa2021second}, and provide decision makers with useful advice, that is methods that can improve trust, enhance user experience, and possibly mitigate the risk of automation bias and deskilling.
\end{enumerate}

Since there is no rose without thorns, here we enumerate the main shortcomings that are associated with a perspectivist approach to supervised ML. 
   \begin{enumerate} 
 \item Need to involve multiple raters: this may represent an important bottleneck in terms of costs or time, and may thus result impractical or expensive in some domain (e.g., medicine) or when dealing with large datasets;
 \item Incompatibility with standard ML approaches, which are usually not designed to take into account multiple perspectives or annotation, and need to design ad-hoc ML methods. While certain classes of learning algorithms (e.g. multi-label ones) may be able to handle multiple labels, it is not clear if these methods can be proficuously applied in the perspectivist ground truthing setting; 
 \item More complex validation/evaluation, due to the absence of a uniquely defined ground truth. While in some cases majority labels can be used as a benchmark ground truth \cite{svenssonautomated2015}, these may not be appropriate in strongly subjective or ambiguous settings.
  \end{enumerate}

\section{Recommendations and a research agenda}
\label{sec:conclusions}

This article aims to disseminate a renovated interest for an alternative approach to ground truthing with respect to the ``reductionist'' one where multiple ratings collected about a single object are reduced into single labels. As we saw in the previous section, Section~\ref{sec:2sides}, this approach entails both advantages and challenges, posed by the will to cope with information richness and manage complexity (also in terms of redundancy, uncertainty and inconsistency) instead of getting rid of it, in light of the research that we presented in Section~\ref{sec:RW}. In this Section, in lieu of a conclusion, we proceed with two sections that shed light on the future: we present a set of agile recommendations for those willing to adopt a perspectivist approach to their ground truthing tasks; and then we propose a number of possible proposals calling for further research and contributions.

\subsection{Recommendations}

In what follows, we share some recommendations to embrace a perspectivist stance in ground truthing. While the impact of some of these practices must still be soundly evaluated, we also mention some of the main studies providing preliminary evidence supporting the recommendations, when available. 

\begin{itemize}
  \item Design annotation schemes that allow raters to associate objects with multiple labels, or also with a 'none of these' label, to account for multiple perspectives directly acknowledged by the single raters. Moreover, allow the raters to express a judgment of inadequacy of the available label set (see \cite{aroyo2015truth});
  \item Involve \emph{enough} raters. This can mean different things depending on the application domain: a number that allows for statistically significant majorities to emerge (e.g., at least 12 raters for dichotomous tasks) or a number of raters that allows to create a majority of superhuman accuracy, according to some estimate of the average accuracy of the raters (see \cite{cabitza2020if,kahneman2016noise});
  \item Involve heterogeneous raters, both in regard to origin and culture and to expertise and skills: different opinions are not always a source of noise, as asserted in~\cite{kahneman2016noise}, but rather of richness \cite{aroyo2015truth,wu2018tagging}.
  
  \item Be aware of the limits of majority voting, especially for complex objects to classify, like in the medical domain. Collecting more information about each judgement, like the rater's confidence, 
  %or what they believe could be the majority choice~\cite{prelec2017solution} 
  should become common practice to help researchers detect these cases, and enable alternative \emph{reductions}~\cite{campagner2021ground} with respect to simple majority, like weighted majority or any strong version of data perspectivism (see Table~\ref{tab:methods}).
  
  \item Evaluate ML models also with respect to \emph{robustness}, or their capability to adequately perform also on external datasets (that is data coming from settings other than where the training data were produced). If performance relevantly degrades on external datasets, this could suggest to adopt a perspectivist approach as a way to mitigate the risk that the models ``overfit'' the ratings of a non-representative sample of users.
  
  \item Report about the rater enrollment process and about the quality of their ratings in detail. %Interested readers can consult the checklist proposed in~\cite{editorial} to report, e.g.,
 In ~\cite{editorial} the authors recommend to describe the process of ground truthing in terms of: a) Number of raters involved to produce the labels; b) The raters' profession and expertise (e.g., years from specialization or graduation); c) The incentive provided, if any; d) Particular instructions given to raters for quality control (e.g., which data were discarded and why); e) how long the labelling process took and, in the case of critical domains such as medicine and law, where and under what conditions it took place (e.g., controlled conditions, real-world interruptions); f) Any chance-adjusted measure of inter-rater agreement (e.g., Rho \cite{cabitza2020if}, Alpha \cite{krippendorff2018content}, Kappa \cite{fleiss1971measuring}); 
 g) The Labelling technique (e.g., majority voting, Delphi method \cite{linstone1975delphi}, consensus iteration).
 
 \item Collect additional information from the raters involved, so as to take actual account of their perspective and way of seeing the objects at hand. For instance, in~\cite{cabitza2020if} we recommended to collect the confidence expressed by the rater about each of their annotations in terms of an ordinal score, as well as other dimensions (e.g., complexity, difficulty, rarity, relevance) along a similar scale. This information can be used to detect the cases for which the raters' confidence was the lowest, thus suggesting that they are the most difficult ones to decide about; or those cases that, for their relevance, it is important that the ML model gets right (i.e., with a sufficient average accuracy) not to mislead its users. %In~\cite{campagnerCDMAKE2021}, we used this information to define a utility metric that is more general, and yet backward compatible, with respect to the metrics that are most commonly used to determine the net benefit of relying on the advice of a specific ML classification model.
 
\end{itemize}
% \begin{enumerate}
% \item	Number of annotators involved to produce the labels;
% \item The annotators' profession and expertise (e.g., years from specialization or graduation);
% \item The incentive provided, if any;
% \item	Particular instructions given to annotators for quality control (e.g., which data were discarded and why);
% \item how long the labelling process took and, in the case of critical domains such as medicine and law, where and under what conditions it took place (e.g., controlled conditions, real-world interruptions);
% \item Any chance-adjusted measure of inter-rater agreement (e.g., Rho \cite{cabitza2020elephant}, Alpha \cite{krippendorff2018content}, Kappa \cite{fleiss1971measuring})
% \item	The Labelling technique (e.g., majority voting, Delphi method \cite{linstone1975delphi}, consensus iteration).
% \end{enumerate}
% \end{itemize}

These recommendations are complementary to those proposed by the \textit{Perspectivist Data Manifesto}\footnote{\url{https://pdai.info}}, a collaborative initiative to promote a perspectivist research agenda in the AI community. In particular, the signatories of the above manifesto also recommend to create and distribute non-aggregated datasets, in order to foster the discussion on the principles of data perspectivism among the research community and to facilitate experimental research in this direction.

\noindent
%Finally, at a more general level, our proposal suggests to rethink any theoretical and experimental research work in predictive computing under the perspectivist lens, asking questions such as \textit{whose opinion is the model relaying in its prediction?} and their implications towards the ethics and responsibility of choices made with the support of predictive models.

% mireille hildebrandt: TROVATA, aggiunta alla fine! ;)

\subsection{Open problems for a possible research agenda}
Finally, in what follows we delineate some possible open problems and research directions that we deem relevant to advance the perspectivist stance in predictive computing:

\begin{enumerate}
\item First, and in connection with the manifesto mentioned in the previous section, we recommend the creation and dissemination of benchmark datasets that could be used to evaluate perspectivist ML models, possibly with respect to different data types (like images, texts, structured data) and for different classification tasks (e.g., detection, risk stratification, forecasting). Such datasets are necessary for the evaluation of novel algorithmic proposals, and they could be used also as benchmarks for setting up challenges, which have recently proven to be important drivers for the development of novel techniques and approaches;

\item Disagreement (and possibly errors) in the multi-rater labels are part and parcel of the perspectivist stance to ground truthing. It is of interest to develop techniques and approaches that are able to exploit the multiple labels to understand and model how the raters err, or on which types of objects they disagree more, so as to develop learning models with the ability to predict the chance that the raters would err on a new object;

\item The usual assumption in ML development is that a model that learnt the majority label from an ensemble of raters could actually provide a better prediction than the majority prediction of an ensemble of models that learnt the single raters' labels. Further research could address this conjecture and investigate if good results (that is useful and reliable predictions) could be achieved also by preserving multiplicity at ground truth level and reducing variability/uncertainty at prediction level through some computational technique. 

\item The perspectivist stance in ground truthing amounts to associating multiple labels to each object. However, especially in low-intersubjectivity tasks, a certain amount of divergent labels is expected in, and may be intrinsic to, the task, and a part of these could be due to errors, inattention or negligence. Thus, it would be interesting to develop ML models that are effectively able to disentangle subjectivity from error, and also to characterize (from a learning theoretic perspective) when this disentanglement is possible at all;

\item In the literature, different algorithmic approaches able to account for a perspectivist ground truth have been proposed (see Table \ref{tab:methods}): these include data augmentation strategies based on the replication of objects associated with multiple labels, ensemble learning methods \cite{campagner2021ground}, or soft loss approaches \cite{uma2020case}. It would be interesting to assess, on real-world problems and applications, the performance of these (and other) approaches, so as to understand their properties and the appropriateness of different algorithms for specific tasks;

\item While some ways to evaluate such ML models have been proposed in the literature (e.g., by evaluating the performance of ML models on objects associated with ``overwhelming majorities'' \cite{campagner2021ground,svenssonautomated2015}, or by adopting soft loss metrics \cite{uma2020case}), these proposals should be unified to develop novel metrics that can better take into account the (chance-adjusted) agreement rate between the raters, or the presence of chance effects and label noise;

\item Similarly to the above point, we need further research on the impact of observer variability on the predictive performance of ML models. Simulating this kind of variability would allow to test for the robustness of these models and get an estimate of the extent the model is ``overfitting'' on either opinions that are not representative of the user population or on partial aspects of the phenomenon of interest.

\item The main concept underlying data perspectivism in ground truthing is that collecting multiple opinions and perceptions about a single phenomenon or object to classify --~and preserving this multiplicity in ML pipelines~-- is valuable, and can convey value to decision makers and users at prediction time through apt decision support \cite{basile2021end,campagner2021ground}. However, a dataset characterized by multiple perspectives, that is by some relevant extent of disagreement between the raters, could be seen as rich as it is noisy. While standard measures of inter-rater agreement (or similar measures, like entropy) can capture this ``noise'' aspect within a \emph{diamond standard}, these measures are not fully capable to represent the informational content and value of such multi-faceted information \cite{campagner2021ground}. Therefore, this approach needs the definition of a theoretical framework (also in terms of the definition of suitable metrics) by which to evaluate the compromise between richness (in that the more different perspectives, the better) and reliability (in that multiplicity does not indicate confusion but complementarity): though we recognize this as a faint approximation of this concept, such an approach could in principle be based on some apt combination of an inter-rater agreement and an information-theoretic (e.g. conditional entropy, or some variants of entropy for imprecise probabilities \cite{yager2018interval}) measure. In doing so, learning-theoretic characterizations of perspectivist ML can be developed, which could enable to understand when a given dataset can be reliable ``ground'' for sound decision support; or, conversely, when its quality needs to be improved with some intervention of \emph{decision hygiene} \cite{kahneman2016noise}, like, e.g., aggregating judgements with methods that leverage professional expertise;

\item As previously described in~\cite{basile2021end,rizos2020average} (see also Section \ref{sec:RW}), the perspectivist stance could also have potential applications and impacts in terms of model interpretability and algorithmic fairness, e.g. by enabling the modeling and detection of biases and discrimination induced by (some groups of) raters. Further research should thus be devoted toward the investigation of possible applications of perspectivist ML in eXplainable and fair AI;

\item Finally, we believe it would be interesting to consider settings in which raters are able to express more information than a single label \cite{aroyo2015truth} – for example, by providing a ranking of the possible labels or expressing their confidence in the labels that they propose~\cite{cabitza2020if}. Given the similarity of these settings to the problems typically investigated in the field of computational social choice \cite{brandt2016handbook} further research should investigate the approaches proposed in that context, and how they could be applied to design perspectivist ML methods that could deal with more general, structured labeling representations.
\end{enumerate}

To conclude, the perspectivist approach is essentially aimed at \emph{caring about} the representativeness and reliability of the ground truth of ML systems. More specifically, and programmatically, data perspectivism fosters wariness toward aggregated gold standards: these reference datasets express single-truth assumptions that can fall short of capturing the necessary complexity of the phenomena for which we want to have support from computational means~\cite{bechmann2019unsupervised}. After all, ML systems are complicated machines that essentially reiterate past judgments and legitimate them~\cite{hildebrandt2020issue} by putting the perceptions of very few (relatively speaking) raters to the attention of countless users and decision makers. There is no guarantee that new objects will be comparable to (or equatable with) those with which such models had been trained, not to speak of the contingent context. We believe that adapting the single-truth assumption of ML 
to the perspectivist paradigm is not only more fair towards minority opinions, but it also has the potential to yield a more accurate and explainable quantitative evaluation of the trained models, as shown by the recent works that are surfacing exploring this research direction~\cite{uma2020case,basile2021end}.
Such a stance creates the necessary room for asking questions such as \textit{whose opinion is the model relying on in its prediction?}, and \textit{what opinions do we want to project into the interpretation of the unexpected new?}; and for reflecting on the implications of these matters on the ethical nature and impact of the decisions made with the support of predictive models.

%\section*{Acknowledgments}
%This work is partially funded by the project ``Be Positive!" (under the 2019 ``Google.org Impact Challenge on Safety” call).


\begin{thebibliography}{10}
  \providecommand{\url}[1]{\texttt{#1}}
  \providecommand{\urlprefix}{URL }
  \providecommand{\doi}[1]{https://doi.org/#1}
  
  \bibitem{akhtar2019polarization}
  Akhtar, S., Basile, V., Patti, V.: A new measure of polarization in the
    annotation of hate speech. In: Alviano, M., Greco, G., Scarcello, F. (eds.)
    AI*IA 2019 -- Advances in Artificial Intelligence. pp. 588--603. Springer
    International Publishing, Cham (2019)
  
  \bibitem{akhtar2020modeling}
  Akhtar, S., Basile, V., Patti, V.: Modeling annotator perspective and polarized
    opinions to improve hate speech detection. In: Proceedings of the AAAI
    Conference on Human Computation and Crowdsourcing. vol.~8, pp. 151--154
    (2020)
  
  \bibitem{aroyo2015truth}
  Aroyo, L., Welty, C.: Truth is a lie: Crowd truth and the seven myths of human
    annotation. AI Magazine  \textbf{36}(1),  15--24 (2015)
  
  \bibitem{artstein2008inter}
  Artstein, R., Poesio, M.: Inter-coder agreement for computational linguistics.
    Computational Linguistics  \textbf{34}(4),  555--596 (2008)
  
  \bibitem{balasubramanian2014conformal}
  Balasubramanian, V., Ho, S.S., Vovk, V.: Conformal prediction for reliable
    machine learning: theory, adaptations and applications. Newnes (2014)
  
  \bibitem{barnard2018inter}
  Barnard, M.E., Pyden, A., Rice, M.S., Linares, M., Tworoger, S.S., Howitt,
    B.E., Meserve, E.E., Hecht, J.L.: Inter-pathologist and pathology report
    agreement for ovarian tumor characteristics in the nurses' health studies.
    Gynecologic oncology  \textbf{150}(3),  521--526 (2018)
  
  \bibitem{basile2021end}
  Basile, V.: It’s the end of the gold standard as we know it. leveraging
    non-aggregated data for better evaluation and explanation of subjective
    tasks. In: Baldoni, M., Bandini, S. (eds.) AIxIA 2020: Advances in Artificial
    Intelligence. XIXth International Conference of the Italian Association for
    Artificial Intelligence, Virtual Event, November 24-27, 2020, Revised and
    Selected papers, vol. 12414. Springer Nature Switzerland AG (2021)
  
  \bibitem{bechmann2019unsupervised}
  Bechmann, A., Bowker, G.C.: Unsupervised by any other name: Hidden layers of
    knowledge production in artificial intelligence on social media. Big Data \&
    Society  \textbf{6}(1),  2053951718819569 (2019)
  
  \bibitem{bender2018data}
  Bender, E.M., Friedman, B.: Data statements for natural language processing:
    Toward mitigating system bias and enabling better science. Transactions of
    the Association for Computational Linguistics  \textbf{6},  587--604 (2018)
  
  \bibitem{beyer2020we}
  Beyer, L., H{\'e}naff, O.J., Kolesnikov, A., Zhai, X., Oord, A.v.d.: Are we
    done with imagenet? arXiv preprint arXiv:2006.07159  (2020)
  
  \bibitem{brandt2016handbook}
  Brandt, F., Conitzer, V., Endriss, U., Lang, J., Procaccia, A.D.: Handbook of
    computational social choice. Cambridge University Press (2016)
  
  \bibitem{editorial}
  Cabitza, F., Campagner, A.: The need to separate the wheat from the chaff in
    medical informatics. International Journal of Medical Informatics
    \textbf{152} (2021)
  
  \bibitem{cabitza2020if}
  Cabitza, F., Campagner, A., Sconfienza, L.M.: As if sand were stone. new
    concepts and metrics to probe the ground on which to build trustable ai. BMC
    Medical Informatics and Decision Making  \textbf{20}(1),  1--21 (2020)
  
  \bibitem{cabitza2019elephant}
  Cabitza, F., Locoro, A., Alderighi, C., Rasoini, R., Compagnone, D., Berjano,
    P.: The elephant in the record: on the multiplicity of data recording work.
    Health informatics journal  \textbf{25}(3),  475--490 (2019)
  
  \bibitem{campagner2021ground}
  Campagner, A., Ciucci, D., Svensson, C.M., Figge, M.T., Cabitza, F.: Ground
    truthing from multi-rater labeling with three-way decision and possibility
    theory. Information Sciences  \textbf{545},  771--790 (2021)
  
  \bibitem{chang2017revolt}
  Chang, J.C., Amershi, S., Kamar, E.: Revolt: Collaborative crowdsourcing for
    labeling machine learning datasets. In: Proceedings of the 2017 CHI
    Conference on Human Factors in Computing Systems. pp. 2334--2346 (2017)
  
  \bibitem{chen2013pairwise}
  Chen, X., Bennett, P.N., Collins-Thompson, K., Horvitz, E.: Pairwise ranking
    aggregation in a crowdsourced setting. In: Proceedings of the sixth ACM
    international conference on Web search and data mining. pp. 193--202 (2013)
  
  \bibitem{cicchetti2006rating}
  Cicchetti, D., Bronen, R., Spencer, S., Haut, S., Berg, A., Oliver, P., Tyrer,
    P.: Rating scales, scales of measurement, issues of reliability: resolving
    some critical issues for clinicians and researchers. The Journal of nervous
    and mental disease  \textbf{194}(8),  557--564 (2006)
  
  \bibitem{dumitrache2018capturing}
  Dumitrache, A., Aroyo, L., Welty, C.: Capturing ambiguity in crowdsourcing
    frame disambiguation. In: Proceedings of the AAAI Conference on Human
    Computation and Crowdsourcing. vol.~6 (2018)
  
  \bibitem{eickhoff2018cognitive}
  Eickhoff, C.: Cognitive biases in crowdsourcing. In: Proceedings of the
    eleventh ACM international conference on web search and data mining. pp.
    162--170 (2018)
  
  \bibitem{eickhoff2013increasing}
  Eickhoff, C., de~Vries, A.P.: Increasing cheat robustness of crowdsourcing
    tasks. Information retrieval  \textbf{16}(2),  121--137 (2013)
  
  \bibitem{fleiss1971measuring}
  Fleiss, J.L.: Measuring nominal scale agreement among many raters.
    Psychological bulletin  \textbf{76}(5), ~378 (1971)
  
  \bibitem{foreman2016generalized}
  Foreman, B., Mahulikar, A., Tadi, P., Claassen, J., Szaflarski, J., Halford,
    J.J., Dean, B.C., Kaplan, P.W., Hirsch, L.J., LaRoche, S., et~al.:
    Generalized periodic discharges and ‘triphasic waves’: a blinded
    evaluation of inter-rater agreement and clinical significance. Clinical
    Neurophysiology  \textbf{127}(2),  1073--1080 (2016)
  
  \bibitem{guan2018said}
  Guan, M., Gulshan, V., Dai, A., Hinton, G.: Who said what: Modeling individual
    labelers improves classification. In: Proceedings of the AAAI Conference on
    Artificial Intelligence. vol.~32 (2018)
  
  \bibitem{heinecke2019crowdpac}
  {Heinecke}, S., {Reyzin}, L.: {Crowdsourced PAC Learning under Classification
    Noise}. In: Proceedings of the Seventh AAAI Conference on Human Computation
    and Crowdsourcing. vol.~7, pp. 41--49. AAAI (2019)
  
  \bibitem{hildebrandt2020issue}
  Hildebrandt, M.: The issue of bias. the framing powers of ml. In: editor, T.
    (ed.) Machine Learning and Society: Impact, Trust, Transparency. MIT Press
    (2021)
  
  \bibitem{jagabathula2017identifying}
  Jagabathula, S., Subramanian, L., Venkataraman, A.: Identifying unreliable and
    adversarial workers in crowdsourced labeling tasks. The Journal of Machine
    Learning Research  \textbf{18}(1),  3233--3299 (2017)
  
  \bibitem{jewett1992potential}
  Jewett, M.A., Bombardier, C., Caron, D., Ryan, M.R., Gray, R.R., Louis, E.L.S.,
    Witchell, S.J., Kumra, S., Psihramis, K.E.: Potential for inter-observer and
    intra-observer variability in x-ray review to establish stone-free rates
    after lithotripsy. The Journal of urology  \textbf{147}(3),  559--562 (1992)
  
  \bibitem{kahneman2016noise}
  Kahneman, D., Rosenfield, A., Gandhi, L., Blaser, T.: Noise: How to overcome
    the high, hidden cost of inconsistent decision making. Harvard Business
    Review  \textbf{94},  38--46 (2016)
  
  \bibitem{kompa2021second}
  Kompa, B., Snoek, J., Beam, A.L.: Second opinion needed: communicating
    uncertainty in medical machine learning. NPJ Digital Medicine  \textbf{4}(1),
    ~1--6 (2021)
  
  \bibitem{krippendorff2018content}
  Krippendorff, K.: Content analysis: An introduction to its methodology. Sage
    publications (2018)
  
  \bibitem{li2018multi}
  Li, S.Y., Jiang, Y., Chawla, N.V., Zhou, Z.H.: Multi-label learning from
    crowds. IEEE Transactions on Knowledge and Data Engineering  \textbf{31}(7),
    1369--1382 (2018)
  
  \bibitem{linstone1975delphi}
  Linstone, H.A., Turoff, M.: The delphi method: Techniques and applications. In:
    The Delphi method: Techniques and applications. Addison-Wesley Publishing
    (1975)
  
  \bibitem{lok1998accuracy}
  Lok, C.E., Morgan, C.D., Ranganathan, N.: The accuracy and interobserver
    agreement in detecting the ‘gallop sounds’ by cardiac auscultation. Chest
     \textbf{114}(5),  1283--1288 (1998)
  
  \bibitem{maul2019intersubjectivity}
  Maul, A., Mari, L., Wilson, M.: Intersubjectivity of measurement across the
    sciences. Measurement  \textbf{131},  764--770 (2019)
  
  \bibitem{mehta2011agreement}
  Mehta, S., Granton, J., Lapinsky, S.E., Newton, G., Bandayrel, K., Little, A.,
    Siau, C., Cook, D.J., Ayers, D., Singer, J., et~al.: Agreement in
    electrocardiogram interpretation in patients with septic shock. Read Online:
    Critical Care Medicine| Society of Critical Care Medicine  \textbf{39}(9),
    2080--2086 (2011)
  
  \bibitem{noble2012minority}
  Noble, J.A.: Minority voices of crowdsourcing: Why we should pay attention to
    every member of the crowd. In: proceedings of the ACM 2012 conference on
    computer supported cooperative work companion. pp. 179--182 (2012)
  
  \bibitem{peterson2019human}
  Peterson, J.C., Battleday, R.M., Griffiths, T.L., Russakovsky, O.: Human
    uncertainty makes classification more robust. In: Proceedings of the IEEE/CVF
    International Conference on Computer Vision. pp. 9617--9626 (2019)
  
  \bibitem{plank2014learning}
  Plank, B., Hovy, D., S{\o}gaard, A.: Learning part-of-speech taggers with
    inter-annotator agreement loss. In: Proceedings of the 14th Conference of the
    European Chapter of the Association for Computational Linguistics. pp.
    742--751 (2014)
  
  \bibitem{rizos2020average}
  Rizos, G., Schuller, B.W.: Average jane, where art thou?--recent avenues in
    efficient machine learning under subjectivity uncertainty. In: International
    Conference on Information Processing and Management of Uncertainty in
    Knowledge-Based Systems. pp. 42--55. Springer (2020)
  
  \bibitem{salminen2018inter}
  Salminen, J.O., Al-Merekhi, H.A., Dey, P., Jansen, B.J.: Inter-rater agreement
    for social computing studies. In: 2018 Fifth International Conference on
    Social Networks Analysis, Management and Security (SNAMS). pp. 80--87. IEEE
    (2018)
  
  \bibitem{schaekermann2018resolvable}
  Schaekermann, M., Goh, J., Larson, K., Law, E.: Resolvable vs. irresolvable
    disagreement: A study on worker deliberation in crowd work. Proceedings of
    the ACM on Human-Computer Interaction  \textbf{2}(CSCW),  1--19 (2018)
  
  \bibitem{sharmanska2016ambiguity}
  Sharmanska, V., Hern{\'a}ndez-Lobato, D., Miguel Hernandez-Lobato, J.,
    Quadrianto, N.: Ambiguity helps: Classification with disagreements in
    crowdsourced annotations. In: Proceedings of the IEEE Conference on Computer
    Vision and Pattern Recognition. pp. 2194--2202 (2016)
  
  \bibitem{sudre2019let}
  Sudre, C.H., Anson, B.G., Ingala, S., Lane, C.D., Jimenez, D., Haider, L.,
    Varsavsky, T., Tanno, R., Smith, L., Ourselin, S., et~al.: Let’s agree to
    disagree: Learning highly debatable multirater labelling. In: International
    Conference on Medical Image Computing and Computer-Assisted Intervention. pp.
    665--673. Springer (2019)
  
  \bibitem{svenssonautomated2015}
  Svensson, C.M., Figge, M.T., Hübler, R.: Automated classification of
    circulating tumor cells and the impact of interobsever variability on
    classifier training and performance. Journal of Immunology Research
    \textbf{2015} (2015)
  
  \bibitem{tao2020label}
  Tao, F., Jiang, L., Li, C.: Label similarity-based weighted soft majority
    voting and pairing for crowdsourcing. Knowledge and Information Systems
    \textbf{62},  2521--2538 (2020)
  
  \bibitem{uma2020case}
  Uma, A., Fornaciari, T., Hovy, D., Paun, S., Plank, B., Poesio, M.: A case for
    soft loss functions. In: Proceedings of the AAAI Conference on Human
    Computation and Crowdsourcing. vol.~8, pp. 173--177 (2020)
  
  \bibitem{vaughan2017making}
  Vaughan, J.W.: Making better use of the crowd: How crowdsourcing can advance
    machine learning research. J. Mach. Learn. Res.  \textbf{18}(1),  7026--7071
    (2017)
  
  \bibitem{wu2018tagging}
  Wu, B., Chen, W., Sun, P., Liu, W., Ghanem, B., Lyu, S.: Tagging like humans:
    Diverse and distinct image annotation. In: Proceedings of the IEEE Conference
    on Computer Vision and Pattern Recognition. pp. 7967--7975 (2018)
  
  \bibitem{yager2018interval}
  Yager, R.R.: Interval valued entropies for dempster--shafer structures.
    Knowledge-Based Systems  \textbf{161},  390--397 (2018)
  
  \bibitem{yun2021re}
  Yun, S., Oh, S.J., Heo, B., Han, D., Choe, J., Chun, S.: Re-labeling imagenet:
    from single to multi-labels, from global to localized labels. arXiv preprint
    arXiv:2101.05022  (2021)
  
  \bibitem{zhang2018ensemble}
  Zhang, J., Wu, M., Sheng, V.S.: Ensemble learning from crowds. IEEE
    Transactions on Knowledge and Data Engineering  \textbf{31}(8),  1506--1519
    (2018)
  
  \end{thebibliography}
\end{document}